\pdfoutput=1
\documentclass[]{fairmeta}
\usepackage{amsthm,amsmath,amssymb}
\usepackage{graphicx}
\usepackage{natbib}
\usepackage{subcaption}
\usepackage{algorithm,algorithmicx,algpseudocode}
\usepackage{pifont} % for \xmark

\usepackage{dblfloatfix}
\usepackage{enumitem}
\usepackage{booktabs}
\usepackage{multirow}
\usepackage{rotating}
\usepackage{bbding}
\usepackage{adjustbox}
\usepackage{xspace}
\usepackage{listings}
\usepackage{xcolor}
\usepackage{colortbl}
\usepackage{xcolor}
\usepackage{wrapfig}
\usepackage{subcaption} 
\definecolor{Red}{RGB}{192, 0, 0}
\definecolor{Blue}{RGB}{12, 114, 186}
\definecolor{Green}{RGB}{120, 190, 50}
\definecolor{Yellow}{RGB}{218, 169, 20}
\definecolor{lightyellow}{RGB}{255,255,153}

\definecolor{HighlightBlue}{RGB}{0, 100, 148}
\definecolor{HighlightRed}{RGB}{230, 57, 70}
\definecolor{myblue}{RGB}{0, 100, 225}

\definecolor{LightRed}{HTML}{ffe0e0}
\definecolor{LightBlue}{HTML}{def5ff}
\definecolor{LightYellow}{HTML}{FFF6DB}
\definecolor{LightGreen}{HTML}{eff9f0}

\newtheorem{theorem}{Theorem}[]

\newtheorem{remark1}[theorem]{Remark}

\newcommand{\OurMethod}{\emph{Point2Insert}}
\newcommand{\Benchmark}{\emph{PointBench}}

\newcommand{\best}[1]{\textcolor{Red}{#1}}
\newcommand{\second}[1]{\textcolor{Blue}{#1}}
\newcommand{\sub}[1]{{\text{#1}}}

\title{Point2Insert: Video Object Insertion via Sparse Point Guidance}

\author[1,2,*]{Yu Zhou}
\author[1,*]{Xiaoyan Yang}
\author[1,3]{Bojia Zi}
\author[1,4]{Lihan Zhang}
\author[1,5]{Ruijie Sun}
\author[2,\dagger]{Weishi Zheng}
\author[1]{Haibin Huang}
\author[1]{Chi Zhang}
\author[1,\dagger]{Xuelong Li}

\affiliation[1]{Institute of Artificial Intelligence, China Telecom (TeleAI)}
\affiliation[2]{Sun Yat-sen University}
\affiliation[3]{The Chinese University of Hong Kong}
\affiliation[4]{Tsinghua University}
\affiliation[5]{Fudan University}

\contribution[*]{Equal Contribution}
\contribution[\dagger]{Corresponding Author}

\begin{document}

\abstract{
This paper introduces \OurMethod, a sparse-point-based framework for flexible and user-friendly object insertion in videos, motivated by the growing popularity of accurate, low-effort object placement. Existing approaches face two major challenges: mask-based insertion methods require labor-intensive mask annotations, while instruction-based methods struggle to place objects at precise locations. \OurMethod\  addresses these issues by requiring only a small number of sparse points instead of dense masks, eliminating the need for tedious mask drawing. Specifically, it supports both positive and negative points to indicate regions that are suitable or unsuitable for insertion, enabling fine-grained spatial control over object locations. The training of \OurMethod\  consists of two stages. In Stage 1, we train an insertion model that generates objects in given regions conditioned on either sparse-point prompts or a binary mask. In Stage 2, we further train the model on paired videos synthesized by an object removal model, adapting it to video insertion. Moreover, motivated by the higher insertion success rate of mask-guided editing, we leverage a mask-guided insertion model as a teacher to distill reliable insertion behavior into the point-guided model. Extensive experiments demonstrate that \OurMethod\  consistently outperforms strong baselines and even surpasses models with $\times$10 more parameters.
}

\maketitle

\begin{figure}[h]
    \centering
    \adjustbox{trim={0.0\width} {0.58\height} {0.12\width} {0}, clip, width=\textwidth}{
        \includegraphics{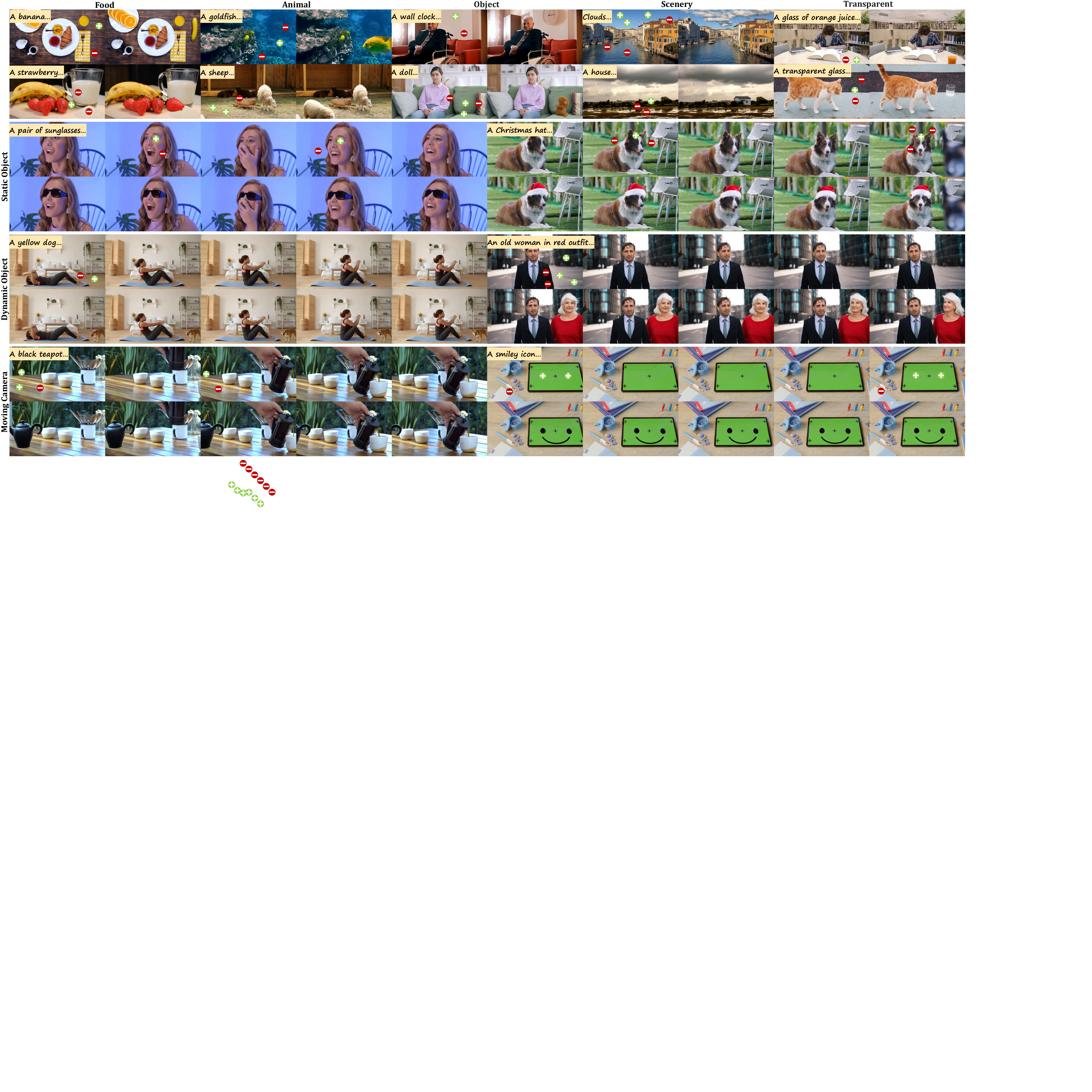} 
      }
      \vspace{-5pt}
      \caption{
        \textbf{We present \OurMethod, a novel sparse point-guided framework designed for precise video object insertion.} 
        By leveraging a small number of sparse \textcolor{Green}{positive}  and \textcolor{Red}{negative} points, our approach bypasses the need for tedious, frame-by-frame masking operations.
        \OurMethod\ handles diverse assets, capable of adding objects from people and animals to household items and background elements. Whether dealing with static or moving targets, or videos with moving cameras, \OurMethod\  consistently produces accurate results with low-effort control points.
      %\noindent \textbf{Project page for this paper is at: } \href{https://zhouyu.github.io/project/x-insert}{\textbf{https://zhouyu.github.io/project/x-insert}}
      }
      \label{fig:teaser}
\end{figure}

\section{Introduction}\label{sec-1-intro}

Video object insertion aims to integrate novel assets into dynamic scenes, serving as a keystone for modern visual effects, commercial content creation, and world model simulation\cite{xing2024survey, melnik2024video, Croitoru_2023, nvidia2025worldsimulationvideofoundation,polyak2025moviegencastmedia}. 
Despite the rapid progress of generative video models~\cite{kling, klingteam2025klingomnitechnicalreport, gen3, sora, hunyuanvideo_kong2024hunyuanvideo, wiedemer2025video}, enabling \textit{flexible} and \textit{user-friendly} insertion with less manual effort remains a challenge.

Current editing methods primarily rely on two types of guidance: text instructions or dense masks. Instruction-based methods offer simplicity but often struggle with accurate object placement\cite{ditto_bai2025scaling, decartai2025lucyedit, wu2025qwenimagetechnicalreport, deng2025bagel, labs2025flux1kontextflowmatching,wang2025ovis,team2025longcat,huang2025diffusion}. 
Conversely, mask-based methods are effective for spatial localization~\cite{bian2025videopainter, vace_jiang2025vace,li2022endtoendframeworkflowguidedvideo}, but the quality of the masks significantly influences the insertion results. Manual mask annotation is time-consuming, especially for long or dynamic sequences. Although segmentation models~\cite{kirillov2023segment, ravi2024sam, carion2025sam3segmentconcepts, chen2017deeplab} can automatically mask objects, they are limited to existing scene entities. When users wish to insert a novel object into an empty region, these models fail to provide position references.  
This motivates a critical question: \textit{Can precise video insertion be achieved through sparse and easy interactions?}

Hence, we leverage points as a flexible and efficient input modality. \textbf{\OurMethod} is a video object insertion framework specifically designed to handle such sparse point inputs. It reformulates the insertion task as a point-guided process, thereby avoiding the need for labor-intensive mask annotations.  A key feature of our approach is its support for both \textbf{positive and negative points}: positive points define the target placement, whereas negative points specify regions to be excluded. This positive–negative point interaction provides users with fine-grained position control over object insertion. 

However, translating such sparse cues into dense video content presents an ill-posed mapping problem.
To bridge the gap between sparse inputs and dense outputs, we develop a two-stage training strategy. In \textit{Stage 1}, we train a foundational insertion model to add objects based on easily constructed
datasets. In \textit{Stage 2}, we fine-tune the model on paired videos generated via an advanced object removal model. 
This dense-to-sparse two stage training paradigm provides a smoother optimization path when adapting the model to sparse control.
Furthermore, we introduce a mask-to-point distillation that transfers reliable insertion behaviors from a mask-guided teacher model to the point-guided model, helping to narrow the gap across different control signals.

We present a data construction pipeline for object-insertion video pairs and curate a large-scale dataset of 1.3M samples from real-world videos. Furthermore, we design \textbf{\Benchmark}, a benchmark featuring multi-dimensional evaluation metrics for insertion performance. Experiments show our method outperforms strong baselines in accurate insertion, including models ten times its size. 
Our method is capable of adding a wide variety of common objects, including food, people, and scenery, demonstrating broad applicability and practicality.

In summary, our contributions are as follows: 

\begin{itemize}[leftmargin=*]
    \item \textbf{First point-guided video insertion framework.} We introduce a sparse interaction paradigm using positive and negative points, significantly reducing annotation overhead while maintaining precise position control.
    
    \item \textbf{Two-stage training with teacher distillation.} 
    We propose a two-stage training pipeline that shifts from dense mask-based to sparse point-based control signals, with additional supervision provided by a mask-guided teacher model.

    \item \textbf{Comprehensive dataset and benchmark.} We construct a point-insertion dataset curated through object removal, along with a \Benchmark\  to evaluate point-based controllability, background preservation, temporal consistency, and visual fidelity.
\end{itemize}

\section{Related Work}\label{sec-2-related_Work}

Recently, numerous video editing methods have been proposed, which can be categorized into two types: training-free and training-based approaches. Training-free methods typically employ DDIM inversion to transform latent into noise and then generate edited videos~\cite{tokenflow_geyer2023tokenflow, videop2p_liu2023video, ku2024anyv2v, tuneavideo_wu2023tune}. These methods eliminate the need for extensive training and large-scale computational resources, but they often suffer from unstable editing results. In contrast, training-based methods require long training times on multiple GPUs~\cite{avid_zhang2023avid,cococo_zi2024cococo,mtv_inpaint_yang2025mtv, revideo_mou2024revideo, vace_jiang2025vace, bian2025videopainter, wu2025insvie, unic, univideo_wei2026univideounifiedunderstandinggeneration,omni_insert_chen2025omniinsertmaskfreevideoinsertion}, yet they produce more reliable and pleasing editing outcomes with lower inference costs. As a result, training-based approaches have become the mainstream in current video editing research~\cite{ma2025step,bao2024vidu,liu2025vfx}. Within this category, a subset of methods leverages masks to guide the editing location, enabling precise modifications while preserving the surrounding content. Other methods rely solely on textual instructions to insert objects into videos, with the editing model itself determining the placement.

\noindent \paragraph{Instruction-based methods.}
Instruction-based video editing has gained increasing popularity in recent years, leading to the emergence of numerous methods. Most existing studies focus on data curation. For instance, InsV2V~\cite{insv2v_cheng2024consistent} employs VideoP2P~\cite{videop2p_liu2023video} to curate training pairs and train editing models. Senorita-2M~\cite{senorita_zi2025se} constructs editing pairs using expert-designed editing strategies. Similarly, InsViE~\cite{wu2025insvie} leverages Stable Video Diffusion~\cite{stability2023svd} to propagate the edited image with the source video, which are subsequently filtered by GPT-4o~\cite{openai2024gpt4technicalreport} and other criteria to retain successful training pairs. Ditto~\cite{ditto_bai2025scaling} provides fewer video pairs but offers higher resolution and longer video durations. OpenVE-3M~\cite{he2025openve} mainly utilizes VACE~\cite{vace_jiang2025vace} and Wan2.1-Fun~\cite{alibaba2023wan2.1fun14bcontrol} to generate large-scale editing pairs, while ReCo~\cite{reco} also releases a video editing dataset containing 500K editing pairs.

In contrast, other methods focus more on architecture design. UNIC~\cite{unic} proposes a model supporting in-context video editing. InstructX~\cite{instructx_mou2025instructxunifiedvisualediting} employs multimodal large language models (MLLMs) to guide the editing process. UniVideo~\cite{univideo_wei2026univideounifiedunderstandinggeneration} and VINO~\cite{vino_chen2026vinounifiedvisualgenerator} unify video generation and editing, enabling object insertion using reference images. VideoCoF~\cite{videocof_yang2025unifiedvideoeditingtemporal} modifies the model to incorporate chain-of-thought reasoning for improved editing performance. ICVE~\cite{icve_liao2025context} trains models on unpaired videos and subsequently performs supervised fine-tuning using filtered data. O-Disco-Edit~\cite{odiscoedit_Chen2025ODisCoEditOD} introduces object distortion control to flexibly support diverse editing cues within a unified representation.
Beyond these general-purpose editors, several methods are specifically designed for object insertion. LoVoRA~\cite{lovora_xiao2025lovoratextguidedmaskfreevideo} jointly addresses object removal and insertion through learnable object-aware localization. Omni-Insert~\cite{omni_insert_chen2025omniinsertmaskfreevideoinsertion} builds a data curation pipeline to enable object insertion from arbitrary reference images. However, due to the difficulty of precisely describing an object's location within a prompt, instruction-based editing methods still struggle to achieve accurate spatial placement in videos.

\paragraph{Mask-based methods.} Video object insertion can be guided by masks. The most straightforward approach is to apply video inpainting methods with given textual prompts. AVID~\cite{avid_zhang2023avid} is the first text-guided video inpainter based on diffusion models, which leverages an any-length technique to enable long video editing. COCOCO~\cite{cococo_zi2024cococo} introduces global damped attention and enhances textual cross-attention to achieve improved consistency and controllability. VideoPainter~\cite{bian2025videopainter} supports any-length video inpainting through a plug-and-play context control block, producing visually pleasing results by leveraging the first edited frame. VideoPivot~\cite{videopivot_xie2025enhancingvideoinpaintingaligned} performs multi-frame consistent image inpainting to generate coherent visual content within masked regions. VACE~\cite{vace_jiang2025vace} is a unified video editing framework capable of inpainting objects in specified masked areas. ReVideo~\cite{revideo_mou2024revideo} utilizes motion information together with spatial masks to guide object inpainting in target regions. Unlike these inpainting methods, VideoAnyDoor~\cite{videoanydoor} is a training-free video object insertion method that achieves high-fidelity detail preservation and precise motion control.
However, when performing object insertion tasks, it is challenging for users to provide accurate frame-by-frame masks. \OurMethod\  allows users to specify the areas where objects should be added by clicking on the target regions, offering a more user-friendly sparse control signal.

\section{Methodology}

\textbf{Overview}. We summarize our approach in two training stages. In Stage 1, we train a model that accepts both masks and point inputs as guidance for object insertion in videos, using an easily constructed dataset. Unlike diffusion-based methods, our Stage‑1 dataset does not rely on generative models; instead, we employ OpenCV with simple yet efficient traditional algorithms to build the dataset~\cite{telea2004image}.
The aim of stage-1 training is to build a model that can be used as the stage-2's foundation, and serve as the stage-2's teacher.  
In Stage 2, we utilize the minimax-remover~\cite{minimax_remover_zi2025minimax} to curate a fine-grained video editing dataset. We then fine-tune the Stage‑1 model with point-based guidance, while leveraging the mask-guided Stage‑1 model as a teacher to further enhance the performance of the point-guided model.

\subsection{Training Dataset Construction}\label{sec-3-3}

% We use  videos from the Internet, resulting in a total of 1.3 million videos.
We curate 1.3M video editing pairs from legally sourced Internet videos using recognition, segmentation, captioning, and object removal. Additionally, in stage-1 training, we used mask-inpainting synthetic data as a proxy for real removals.

\subsubsection{Video Object Segmentation and Captioning}  
We detect objects in the first frame using Rex-Omni~\cite{rexomni} and segment them with SAM2~\cite{sam2_ravi2025sam2} to obtain video masks. Objects of extreme scales (i.e., occupying more than 50\% or less than 0.5\% of the frame area) are discarded. To balance object categories, we constrain the fraction of videos from the top 1\% classes to at most 1\%. Furthermore, we employ Qwen-VL2.5-7B~\cite{qwen2.5-VL} to caption the videos, generating both (1) global captions that describe scene context and spatial relations, and (2) object-level captions that capture appearance and pose. During training, we sample object captions, global captions, and class names in a 90:5:5 ratio to improve robustness.

% \begin{figure}[t]
%   \centering
%   % \vspace{-12pt}
%   \scalebox{1.5}{
%       \adjustbox{trim={0.0\width} {0.897\height} {0.768\width} {0}, clip, width=0.5\textwidth}{
%         \includegraphics{fig/data-pipe.pdf}
%       }
%   }
%   \vspace{-10pt}
%   \caption{
%    \textbf{Training Dataset Construction.} Our pipeline detects objects within videos, extracts segmentation masks, removes the objects, and extracts video captions. Finally, we sample positive and negative points for training.
%    }
%   \vspace{-2pt}
%   \label{fig:datapipe}
% \end{figure}

\begin{figure}[t]
  \centering
  \begin{subfigure}[b]{0.55\textwidth}
    \centering
    \adjustbox{trim={0.0\width} {0.897\height} {0.768\width} {0}, clip, width=\textwidth}{
      \includegraphics{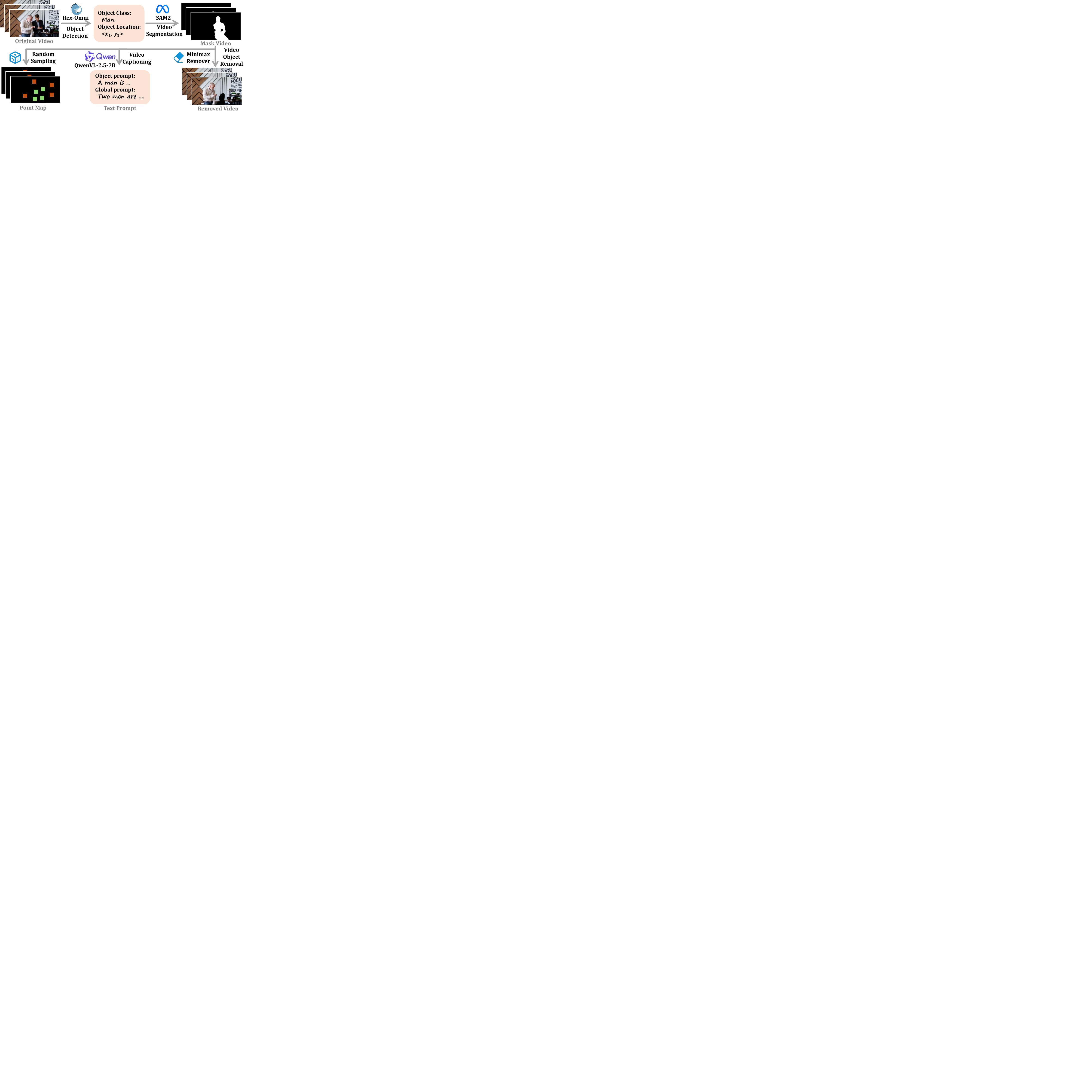}
    }
    \vspace{-2pt}
    \caption{
      \textbf{Training Dataset Construction.} 
    }
    \label{fig:datapipe}
  \end{subfigure}
  \hfill
  \begin{subfigure}[b]{0.40\textwidth}
    \centering
    \adjustbox{trim={0.0\width} {0.737\height} {0.75\width} {0}, clip, width=\textwidth}{
      \includegraphics{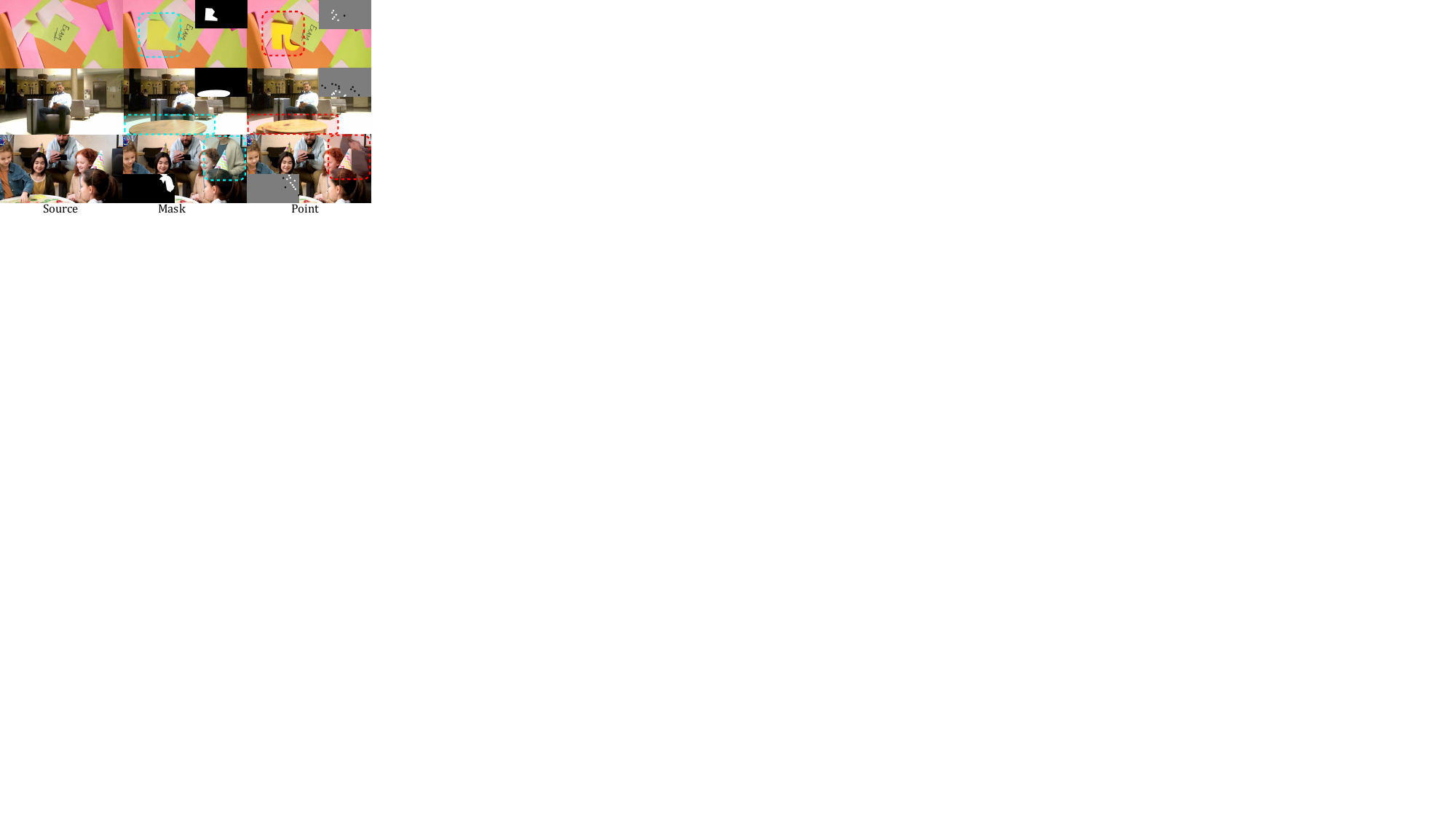}
    }
    \vspace{-2pt}
    \caption{
      \textbf{Comparison of mask- and point-only results.} 
    }
    \label{fig:mask-vs-point}
  \end{subfigure}
  \vspace{-2pt}
  \caption{
    \textbf{(a)Training Dataset Construction}:
    Our pipeline detects objects within videos, extracts segmentation masks, removes the objects, and extracts video captions. Finally, we sample positive and negative points for training.
    \textbf{(b)Comparison of mask- and point-only results}:
    The mask-based setting generates more natural-looking objects due to explicit boundary cues. Point maps without clear boundaries often result in irregular shapes, geometric distortions, and blurred appearances.
  }
  \label{fig:combined}
\end{figure}

\subsubsection{Stage-1 Training Dataset Construction} 
The Stage-1 dataset is straightforward to construct. In this stage, the model receives two types of positional guidance (i.e., masks and points), corresponding to the masked video and the inpainted video, denoted as $\mathbf{x}_\sub{m}, \textbf{x}_\sub{inp} \in \mathbb{R}^{f \times h \times w \times c}$, where $\mathbf{x}_\sub{m} = \textbf{x} \odot (\textbf 1-\textbf{m})$, $\mathbf{m}$ is the binary mask and $\mathbf{x}$ is the original video. Importantly, the inpainted video is not generated by diffusion models; instead, it is produced using a traditional inpainting algorithm~\cite{telea2004image}. To unify the two types of position guidance, we convert the points into the point map, denoted as $\mathbf{x}_\sub{p} \in \mathbb{R}^{f \times h \times w \times c}$, having the same dimension as the mask $\mathbf{m}$. In particular, we extend several pixels from the point center to the surrounding area, thereby constructing a square region. Within this region, positive points are assigned a value of 1, negative points are assigned a value of 0.5, and the background is set to 0.
As the construction process is computationally efficient, these videos are generated online during training.

\subsubsection{Stage-2 Training Dataset Construction}
The automated pipeline is shown in Fig.~\ref{fig:combined}~\subref{fig:datapipe} . We leverage SOTA object remover to synthesize training pairs $(\mathbf{x}_\sub{src}, \textbf{x})$ by removing objects from original videos $\mathbf{x} \in \mathbb{R}^{f\times h \times w \times c}$, effectively treating insertion as the inverse of removal. The point map curation is same as the stage-1.

\begin{figure*}[t]
  \centering
  % \adjustbox{trim={0.0\width} {0.883\height} {0.572\width} {0}, clip, width=\textwidth}{
  %   \includegraphics{fig/overview.pdf} 
  % }
    \includegraphics[width=.95\textwidth]{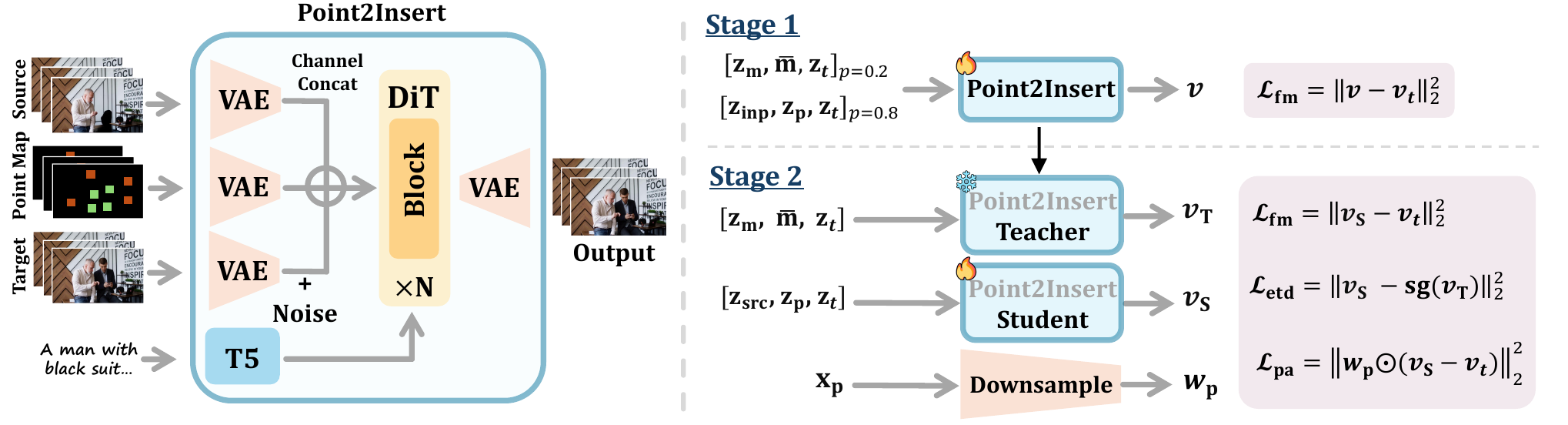} 

  \vspace{-2pt}
  \caption{
   \textbf{Model overview.} 
   \textbf{The left panel} illustrates that \OurMethod\  maps the source, point map, and target videos into a latent space via a VAE. These latents are concatenated channel-wise and processed by a DiT-based denoiser.
   \textbf{The right panel} details our two-stage training strategy:
   \textit{Stage 1: Insertion Model Pre-training.} We use traditional lightweight inpainting to create source videos with objects removed. The DiT is then trained using a flow matching objective with maps generated via hybrid mask and point sampling.
    \textit{Stage 2:  Stronger Model Distillation.} A frozen teacher model from Stage 1 supervises a student model. While the teacher uses masks, the student employs point maps and source videos from a removal model. For further refinement, weight maps downsampled from the point-based map are applied to a weighted flow matching loss.
}
  \vspace{-2pt}
  \label{fig:overview}
\end{figure*}

\subsection{Stage-1 Training: Pretrain An Insertion Model}

The objective of stage-1 training is to build a foundation model for stage-2 and to serve as a teacher that improves the performance of the point-guided video editor in stage-2. To achieve this, we use masked videos $\mathbf{x}_\sub{m}$ together with masks $\mathbf{m}$ to train the mask-guided video insertion model, and employ inpainting videos $\mathbf{x}_\sub{inp}$ along with point maps $\mathbf{x}_\sub{p}$ to train the foundation model. Since both types of inputs share the same dimensionality, we unify them into a single model with a unified training process. During inference, the model determines the editing task based on whether the input video contains corrupted regions, eliminating the need for additional control signals to distinguish between the two editing tasks.

During training, we employ the autoencoder $\mathcal{E}$ to encode the video conditions into latent representations. Specifically, the masked video $\mathbf{x}_\sub{m}$ is encoded as $\mathbf{z}_\sub{m}$, and the corresponding mask is encoded as $\mathbf{\bar{m}}$. The original video $\mathbf{x}$ is encoded into $\mathbf{z}$. Similarly, the inpainted video $\mathbf{x}_\sub{inp}$ and the point map $\mathbf{x}_\sub{p}$ are encoded into $\mathbf{z}_\sub{inp}$ and $\mathbf{z}_\sub{p}$, respectively. Given a timestep $t \in [0,1]$, the noisy latent is constructed as $\mathbf{z}_t = t \cdot \epsilon + (1 - t) \cdot \mathbf z$, where $\epsilon \sim \mathcal{N}(0,I)$,
following the flow matching training strategy proposed in~\cite{flow_match_lipman2023flow, lipman2024flow}. 

For the mask-guided task, we concatenate $\mathbf {z}_\sub{m}$, $\bar{\mathbf m}$, and $\mathbf {z}_{t}$ along the channel dimension and feed the resulting representation into the DiT model with a probability of $p=0.2$, while for the point-guided task, we concatenate $\mathbf z_{\sub{inp}}$, $\mathbf {z}_\sub{p}$, and $\mathbf z_t$ in the same manner and input them into the DiT model with a probability of ${p}=0.8$. To accommodate these heterogeneous inputs, we replace the original patch embedder of the DiT model with a new patch embedder that supports 48 input channels.

\subsection{Stage-2 Training: Distill An Stronger Insertion Model}\label{sec-3-2}

While point-based interaction offers great convenience for users, training a robust model under such sparse conditions face a challenge. There is a significant performance gap exists between discrete points and the dense masks used in object insertion, as shown in Fig.~\ref{fig:combined}~\subref{fig:mask-vs-point}. Intuitively, denser information helps the model produces more consistent motion and precise location. Therefore, only using sparse signal to guide the insertion model's training often leads to slow convergence and suboptimal performance. 
To bridge this gap, we propose a distillation strategy~\cite{hinton2015distilling} by finetuning the stage-1 model with points guidance supervised by mask-guided teacher.

% \begin{wrapfigure}{l}{0.45\textwidth} 
%   \centering
%   % \vspace{-12pt}
%   \scalebox{0.9}{
%       \adjustbox{trim={0.0\width} {0.737\height} {0.75\width} {0}, clip, width=0.46\textwidth}{
%         \includegraphics{fig/point-vs-mask.pdf}
%       }
%   }
%   % \vspace{-6pt}
%   \caption{
%    \textbf{Comparison of mask-only and point-only results at the same training steps.} 
%     The mask-based setting is able to generate more natural-looking objects, owing to the explicit boundary provided by the mask. In contrast, point maps without clear boundary cues often result in objects with irregular boundaries, geometric distortions, and blurred appearances.
%    }
%   % \vspace{-8pt}
%   \label{fig:mask-vs-point}
% \end{wrapfigure}

\textbf{Explicit Teacher Distillation.}
Here, we introduce 
explicit teacher distillation to achieve goal of the generation consistency between point-based and mask-based model. The following is our distillation loss:

\begin{equation}
\begin{aligned}
    \mathcal{L}_{\text{etd}} &= \left\| \mathbf{v_{\text{S}}} - \text{sg}(\mathbf{v_{\text{T}}}) \right\|_2^2, \\
    \mathbf{v_{\text{S}}} &= u_{\text{student}}(\mathbf{z_{\text{src}}, z_{\text{p}}}, \textbf{z}_t, \textbf{c}_\text{txt}, t; \theta_\text{S}), \\
    \mathbf{v}_{\text{T}} &= u_{\text{teacher}}(\mathbf{z}_\text{m}, \bar{\textbf{m}}, \textbf{z}_t, \textbf{c}_\text{txt}, t; \theta_\text{T}).
\end{aligned}
\end{equation}

\hspace{-6mm}where, $\text{sg}(\cdot)$ is the stop gradient operation, the $\theta_\text{S}$ and $\theta_\text{T}$ are the parameters of the student $u_\text{student}$ and teacher model $u_\text{teacher}$, respectively. $\mathbf{c}_\text{txt}$ is the text condition encoded by T5~\cite{t5_raffel2020exploring}. The $\mathbf{v}_\text{S}$ and $\mathbf{v}_\text{T}$ are the output of the student and teacher model, while $\mathbf{v}_{t}$ is the velocity: $\mathbf{v}_{t} = \epsilon - \mathbf{z}$.

By enforcing output consistency across different guidance types, distillation bridges the performance gap between dense masks and sparse points.

\textbf{Point-Aware Enhancement.}
Since control points occupy only a negligible fraction of the pixel space, we introduce a point-aware enhancement loss to address signal sparsity. 
We transform the point map $\mathbf{x}_\text{p}$ into weight map $\mathbf{w}_\text{p}$ to reweight the distillation loss, aiming to force model focus on the small but critical regions.

To ensure the point map $\mathbf{x}_\text{p} \in \mathbb{R}^{f \times h \times w \times c}$ matches the dimensions of the target latent $\mathbf{z}_t \in \mathbb{R}^{(\frac{f-1}{4}+1) \times \frac{h}{8} \times \frac{w}{8} \times c'}$, where $c'=16$ in current mainstream video diffusion models. we employ a spatial-temporal average pooling operator $\mathcal{T}$. The strides are $1 \times 8 \times 8$ for the first frame and $4 \times 8 \times 8$ for subsequent frames. We then repeat the first channel with $c'$ times to align the weight channels with the latent space.
Unlike direct downsampling by the VAE encoder, the pooling operation maps the source point map to the weight map in a continuous manner, assigning higher weights to regions with higher point density.

Consequently, given the ground-truth velocity $\textbf{v}_t=\epsilon-\textbf z$, the loss $\mathcal{L}_{\textrm{pa}}$ is defined as:

\begin{equation}
\begin{aligned}
    \mathcal{L}_{\textrm{pa}} 
&= \left\| \mathbf{w}_\text{p} \odot \left(\mathbf{v_{\text{S}} - v}_t \right)\right\|_2^2, \\
   \textbf{w}_\text{p} &= \left| \mathcal{T}(\mathbf{x}_\text{p}) - 0.5 \right|.
\end{aligned}
\end{equation}

This weighting strategy encourages model to focus on annotated locations, ensuring modification on editable regions and preservation on uneditable regions, respectively.

\textbf{Background Alignment.}
However, the current video object remover is not fully reliable, as it occasionally generates unacceptable results. These flawed cases within the training dataset can adversely affect the learning process. The majority of failures stem from background inconsistencies, which cause the editing model to produce incoherent backgrounds and thereby degrade the overall generation quality. To address this issue, we apply dilation and feathering to the point map, followed by compositing the input video’s background onto the target frames. This approach ensures consistent background reconstruction while effectively eliminating potential boundary artifacts.

\textbf{Training Objective.}
The final training loss $\mathcal{L}$ is defined as a weighted combination of the control and generative terms:
\begin{equation}
\mathcal{L} = \mathcal L_{\textrm{fm}} + \lambda_1\mathcal{L}_{\text{etd}} + \lambda_2 \mathcal{L}_{\text{pa}},
\end{equation}
where coefficients $\lambda_1$ and $\lambda_2$ weight the trade-off between distillation guidance and sparse point control. The flow-matching loss $\mathcal{L}_{\mathrm{fm}}$ is defined as the discrepancy between the student output $\mathbf{v_{\text{S}}}$ and the ground-truth target $\mathbf{v}_{t}$.\\

\section{Experiments}\label{sec-4-exp}

\subsection{Implementation Details}

\paragraph{\textbf{Training Details.}}

Our method is built upon the WAN2.1-T2V-1.3B~\cite{wan2025}.
To accommodate varying resolutions, 
we employ a dynamic resolution strategy using buckets ranging 
from $240 \times 240 \times 1$ to $832 \times 832 \times 121$, 
with an interval of $16 \times 16 \times 4$. 
The training process consists of two stages, both optimized using AdamW~\cite{adamw_loshchilov2017decoupled} ($\beta_1 = 0.9, \beta_2 = 0.99$). 
\textbf{Stage I.} We train the model for 5,000 steps on a mixture of masked (80\%) and inpainted (20\%) video data. This stage uses a batch size of 128 and a learning rate of $5 \times 10^{-5}$.
\textbf{Stage II.}
We further fine-tune the model for 5,000 steps on object removal tasks with a reduced learning rate of $1 \times 10^{-5}$. We set loss weights $\lambda_1 = 1.5$ and $\lambda_2 = 1.2$. To enhance insertion across different point density, we sample insertion prompts according to the following distribution: masks (10\%), sparse points (30\%), and dense points (60\%).

% \vspace{-5pt}
\paragraph{\textbf{Benchmark.}}
To evaluate our point-based insertion model, we establish \Benchmark, which consists of two specialized subsets:
\begin{itemize}[leftmargin=*,itemsep=-0.1em]
\item \textit{Point-based Control:} We utilize the DAVIS testset~\cite{perazzi2016benchmark}, which contains 90 videos characterized by rapid motion, significant blur, and complex occlusions. For each video, we sample keyframes at 10-frame intervals and manually annotate positive and negative points, alongside descriptive text for the target objects.
\item \textit{Mask-based Control:} We collect 100 diverse web videos covering both indoor and outdoor scenery. Leveraging Qwen-VL~\cite{qwen2.5-VL} and SAM2~\cite{sam2_ravi2025sam2}, we implement an automated pipeline to extract object-level prompts and masks to construct evaluation pairs.
\end{itemize}

\paragraph{\textbf{Metrics.}} 
We evaluate 11 metrics across four dimensions:

\begin{itemize}[leftmargin=*,itemsep=-0.1em]

\item[$\bullet$] \textit{Point Response Accuracy.} 
We employ Sa2va~\cite{perazzi2016benchmark} to segment the newly added objects and calculate the hit rate for positive and negative points within the predicted masks.

\item[$\bullet$] \textit{Background Preservation.} We compute PSNR~\cite{psnr}, LPIPS~\cite{lpips}, SSIM~\cite{ssim}, MSE, and MAE specifically within the unmasked regions to assess reconstruction quality.

\item[$\bullet$] \textit{Text Alignment.} We evaluate text-video alignment (CLIP-TA) by comparing the generated frames with the object insertion prompts. Additionally, background semantic preservation (CLIP-BG) is measured by comparing the background regions against the global scene captions.

\item[$\bullet$] \textit{Temporal Consistency.} We utilize Ewarp (pixel-level) and CLIP-TC (semantic-level) to quantify the discrepancy between consecutive frames, with results averaged across the entire video sequence.

\end{itemize}

\paragraph{\textbf{Baselines.}}
We evaluate \OurMethod\  against several state-of-the-art video editing methods, categorized into two groups: 
(i) \textbf{mask-based} methods, including VideoPainter~\cite{bian2025videopainter}, Vace~\cite{vace_jiang2025vace}, and Senorita~\cite{senorita_zi2025se}; 
and (ii) \textbf{instruction-based} approaches, such as Ditto~\cite{ditto_bai2025scaling}, Lucy~\cite{decartai2025lucyedit}, VideoCof~\cite{videocof_yang2025unifiedvideoeditingtemporal}, ICVE~\cite{icve_liao2025context}, and UniVideo~\cite{univideo_wei2026univideounifiedunderstandinggeneration}.
Although Senorita is a mask-free video propagation model, it requires a mask to edit the first frame. 
Since some baselines do not natively support point-based or mask-based control, we adapt their inputs to ensure a fair comparison:
\begin{itemize}[leftmargin=*,itemsep=-0.1em]
\item \textit{Instruction-based models:} To convert point or mask inputs into textual prompts, we employ Qwen2.5-VL-7B~\cite{qwen2.5-VL} to translate these annotations into text descriptions that specify the target locations.

\item \textit{Mask-based models:} To convert point inputs into masks, we compute the convex hull of the input points on each keyframe and apply the resulting masks consistently across the entire video.
\end{itemize}

% \vspace{-2pt}
\subsection{Experimental Results}

\begin{table*}[t]
\centering
\small
\caption{\textbf{Quantitative comparison of \OurMethod and video editing models on \Benchmark}. Evaluation consists of two parts:
(1) using point maps as input on the DAVIS dataset, and
(2) using mask as input on an internal test dataset.
We compare against both instruction-based models: 
Ditto~\cite{ditto_bai2025scaling}, Lucy~\cite{decartai2025lucyedit}, ICVE~\cite{icve_liao2025context}, Senorita~\cite{senorita_zi2025se}, UniVideo~\cite{univideo_wei2026univideounifiedunderstandinggeneration} and  VideoCoF~\cite{univideo_wei2026univideounifiedunderstandinggeneration}, and mask-based models: VideoPainter\cite{bian2025videopainter} and VACE\cite{vace_jiang2025vace}.
Evaluation metrics include success rate in responding to the provided control conditions,
Background preservation in unedited regions,
Text-alignment quality, and
Inter-frame consistency.
\best{Red} indicates the best, and \second{blue} denotes the second.
Our model has only \textbf{1.3B} parameters, yet it outperforms other models that are \textbf{10$\times$} larger in scale. Model parameter details are provided in Appendix~\cref{supp:param}.
}
\vspace{-0pt}
\scalebox{0.78}{
\setlength{\tabcolsep}{1.2mm}{
\begin{tabular}{cl|cc|ccccc|cc|cc}
\toprule
    \multicolumn{2}{c|}{\multirow{1}{*}{\textbf{Point-Based Eval}}} 
    & $\textbf{Acc}_{\textbf{pos}}\uparrow$ 
    & $\textbf{Acc}_{\textbf{neg}}\uparrow$
    & $\textbf{MSE}\downarrow$ 
    & $\textbf{MAE}\downarrow$ 
    & $\textbf{PSNR}\uparrow$ 
    & $\textbf{SSIM}\uparrow$ 
    & $\textbf{LPIPS}_{\times 100}$ $\downarrow$ 
    & $\textbf{CLIP BG}\uparrow$ 
    & $\textbf{CLIP TA}\uparrow$ 
    & $\textbf{CLIP TC}\uparrow$ 
    & ${\textbf{Ewarp}_{\times 100}}$ $\downarrow$ 
     \\

\midrule

\multirow{5}{*}{\rotatebox{90}{\textbf{Instruction.}}} & 
\textbf{Ditto}    
    & 22.07 & 81.99 & 5179.42 & 53.47 & 12.28 & 0.4298 & 36.33 & 21.95 & 22.01  & \best{97.41} & 7.19  \\
    
& \textbf{Lucy} 
    & 16.19 & 89.29 & 449.69 & 11.70 & 23.09 & 0.7441 & 13.32 & 24.20 & 22.92 &  96.21 &  6.97  \\

& \textbf{ICVE}    & 14.65  & 90.39 & 848.60     & 15.39     & 23.23 & 0.7328 & 12.69 & 24.48 & 22.99 & 95.91 & 9.32  \\
& \textbf{UniVideo} & 15.73  & 84.87  & 2419.20 & 29.22 & 16.29 & 0.5054 & 24.34 & 24.08 & 23.39 & 96.82 & 6.45 \\
& \textbf{VideoCoF} & 8.15  & 93.51 & 341.80 & 12.20 & 23.95 & 0.8184 & 7.22 & 24.05 & 22.42 & 96.27 & \best{5.77}  \\

\midrule

\multirow{3}{*}{\rotatebox{90}{\textbf{Mask.}}}
& \textbf{Senorita} 
    & 49.63  & {94.73}  & 225.28 & 6.98  & 26.94 & 0.8201 & 7.29 & 24.85 & \second{23.66} & 96.76 &  \second{6.02} \\
& \textbf{VideoPainter} 
    &  \second{68.36} & 88.70& 226.38 & 9.02 & 25.50 & {0.8142} & 8.00 & \second{24.99} & {23.36} & {96.49}& {6.31} \\
& \textbf{Vace}         
    & 16.04 & \second{94.87} & \second{140.88} & \second{6.40} & \second{28.22} & \second{0.8444} & \second{5.21} & 24.74 & 22.74 & 96.82 &  6.28  \\

\midrule

& \textbf{Ours} 
& \best{81.13} & \best{96.87} & \best{93.19} & \best{5.39}  & \best{29.51}  & \best{0.8592}  & \best{4.49}  & \best{25.55} & \best{23.68} & \second{96.86}  &  6.25  \\

\bottomrule
\toprule

\multicolumn{2}{c|}{\multirow{1}{*}{\textbf{Mask-Based Eval}}} 
& 
\multicolumn{2}{|c|}{\textbf{Model Size}}
& $\textbf{MSE}\downarrow$ 
& $\textbf{MAE}\downarrow$ 
& $\textbf{PSNR}\uparrow$ 
& $\textbf{SSIM}\uparrow$ 
& $\textbf{LPIPS}_{\times 100}\downarrow$ 
& $\textbf{CLIP BG}\uparrow$ 
& $\textbf{CLIP TA}\uparrow$ 
& $\textbf{CLIP TC}\uparrow$ 
& $\textbf{Ewarp}_{\times 100}\downarrow$ 
 \\

\midrule

\multirow{5}{*}{\rotatebox{90}{\textbf{Instruction.}}} 
& \textbf{Ditto}    
&  \multicolumn{2}{|c|}{14B} & 7877.31  &  70.03  &  12.07  &  0.6286  &  37.60  &  16.82  &  17.53  &  \best{99.32} & 0.41 \\

& \textbf{Lucy} 
&  \multicolumn{2}{|c|}{5B} & 662.92  &  9.18  &  27.85  &  0.9017  &  7.66  &  23.15  &  21.74  &  99.08 &  0.57 \\

&\textbf{ICVE} 
&  \multicolumn{2}{|c|}{13B} & 2998.15   & 35.64 & 20.10 & 0.7993 & 14.06 & 23.83 & 22.32 & 99.20 & 0.74 \\
&\textbf{UniVideo} 
&  \multicolumn{2}{|c|}{13B+7B}  & 813.82  & 13.94 & 21.27 & 0.7320 & 10.92 & 23.76 & 22.66 & 99.18 & 0.37 \\   % Qwen VL + hunyuan T2V
&\textbf{VideoCoF} 
&  \multicolumn{2}{|c|}{14B} & 442.63  & 9.73  & 25.58 & 0.9171 & 7.23 & 23.38 & 22.06 & 99.15 & \best{0.29} \\

\midrule

\multirow{3}{*}{\rotatebox{90}{\textbf{Mask.}}} 
&\textbf{Senorita} 
&  \multicolumn{2}{|c|}{5B+12B} & \second{106.35 } & \second{4.78}  & 30.91 & {0.9445} & 4.25 & \best{24.66} & 22.91 & \second{99.31} & 0.34 \\
& \textbf{VideoPainter} 
&  \multicolumn{2}{|c|}{5B+12B} & 205.24 & 4.31 & 29.24 & 0.9520 & 3.84 & 24.52 & \best{23.14} & 99.26 & 0.39\\    % ?
& \textbf{Vace}         
&  \multicolumn{2}{|c|}{1.3B+0.3B} & 164.84  &  3.74  &  \second{32.18}  &  \second{0.9566}  &  \second{3.25}  &  24.32  &  22.73  &  99.11   
& 0.49 \\

\midrule

& \textbf{Ours} 
&  \multicolumn{2}{|c|}{\textbf{1.3B}} & \best{103.09}  &  \best{2.59}  &  \best{35.72}  &  \best{0.9694}  &  \best{2.02}  &  \second{24.53}  &  \second{23.11}  &  {99.27} & \second{0.32} \\

\bottomrule
\end{tabular}
}}
\label{tab:benchmark}
\end{table*}

\begin{table}[t]
\centering
\small
\caption{
\textbf{Quantitative Evaluation of Generation Quality.}  
We evaluate the generation quality of methods on \Benchmark\  using a large multimodal language model (MLLM).
The MLLM assesses object insertion success rate (\textbf{Succ.}),
semantic alignment between the generated 
object and the textual description (\textbf{Sem.}),
and the visual quality of the generated object (\textbf{Qual.}).
}
\scalebox{0.95}{
\setlength{\tabcolsep}{2.5mm}{
\begin{tabular}{l|ccc|ccc}
\toprule
\multirow{2}{*}{\textbf{Method}} 
 & \multicolumn{3}{c|}{\textbf{GPT5.2}} 
 & \multicolumn{3}{c}{\textbf{Gemini3Pro}} \\
\cmidrule(lr){2-4} \cmidrule(lr){5-7}
& \textbf{Succ.}(\%)$\uparrow$ 
& \textbf{Sem.}$\uparrow$ 
& \textbf{Qual.}$\uparrow$ 
& \textbf{Succ.}(\%)$\uparrow$ 
& \textbf{Sem.}$\uparrow$ 
& \textbf{Qual.}$\uparrow$ \\
\midrule
\textbf{Lucy}     & 28.89 & 1.42 & 1.55 & 21.11 & 1.50 & 1.45 \\
\textbf{Ditto}    & 35.56 & 1.80 & 2.03 & 25.56 & 1.69 & 1.45 \\
\textbf{ICVE}      & 20.00 & 1.39 & 1.57 & 20.00    & 1.61 & 1.58 \\
\textbf{UniVideo}  & 36.67 & 1.80  & 1.97 & 24.44 & 1.87 & 1.77 \\
\textbf{VideoCof}  & 10.00 & 1.14 & 1.32 & 7.78  & 1.14 & 1.14 \\
\midrule
\textbf{Senorita}  & 56.18 & 1.68 & 2.04 & {56.18} & 2.23 & 1.90 \\
\textbf{Video Painter}      & \second{82.02} & \second{2.01} & \second{2.31} & \second{76.40} & \second{2.64} & \second{2.14} \\
\textbf{Vace}     & 18.89 & 1.34 & 1.59 & 20.00 & 1.42 & 1.50 \\

\midrule
\textbf{Ours} & \best{91.11} & \best{2.52} & \best{2.83} & \best{94.44} & \best{3.69} & \best{3.57} \\
\bottomrule
\end{tabular}
}}
\label{tab:mllm}
\end{table}

\paragraph{\textbf{Quantitative Comparisons.}}
Tab.~\ref{tab:benchmark} presents the quantitative comparisons on \Benchmark.
\OurMethod\  demonstrates strong performance under sparse point control. The upper part of Tab.~\ref{tab:benchmark} presents the quantitative comparison in point setting. 
Specifically, our method performs best in control accuracy, outperforms the second-best method by 12.77\% and 8.17\% in $\text{Acc}_\text{pos}$ and $\text{Acc}_\text{neg}$, indicating our ability to accurately respond to the point control and insert objects that satisfy both positive and negative constraints. 

In terms of background preservation, \OurMethod\  ranks first in all five metrics, capability to avoid background interference while performing local additions. 
In contrast, methods such as UniVideo and Ditto suffer from noticeable performance degradation due to interference with the overall video style.
Additionally, \OurMethod\  maintains competitive performance in text alignment and temporal consistency, including CLIP-BG, CLIP-TA, and CLIP-TC. Overall, these results show that \OurMethod\  effectively adds objects under sparse control, successfully balancing control accuracy, generation quality, background preservation, and temporal consistency.

In the mask-based editing task, \OurMethod\  also outperforms most baselines. It leads the second-best VACE by 3.54 in PSNR, by 1.23 in LPIPS, and exhibits superior background preservation. In the only metric where we are not in the top two, CLIP-TC, we are just 0.05\% behind the top performer. This demonstrates that our model performs equally well on mask-based editing tasks. The consistently strong performance across both point- and mask-based tasks highlights the effectiveness of our method in handling both sparse and dense control signals.

% \vspace{-5pt}

\begin{figure*}[htbp]
  \centering
  % \adjustbox{trim={0.0\width} {0.627\height} {0.656\width} {0}, clip, width=\textwidth}{
  %   \includegraphics{fig/comparison.pdf} 
  % }
  \adjustbox{trim={0.0\width} {0.565\height} {0.656\width} {0}, clip, width=0.93\textwidth}{
    \includegraphics{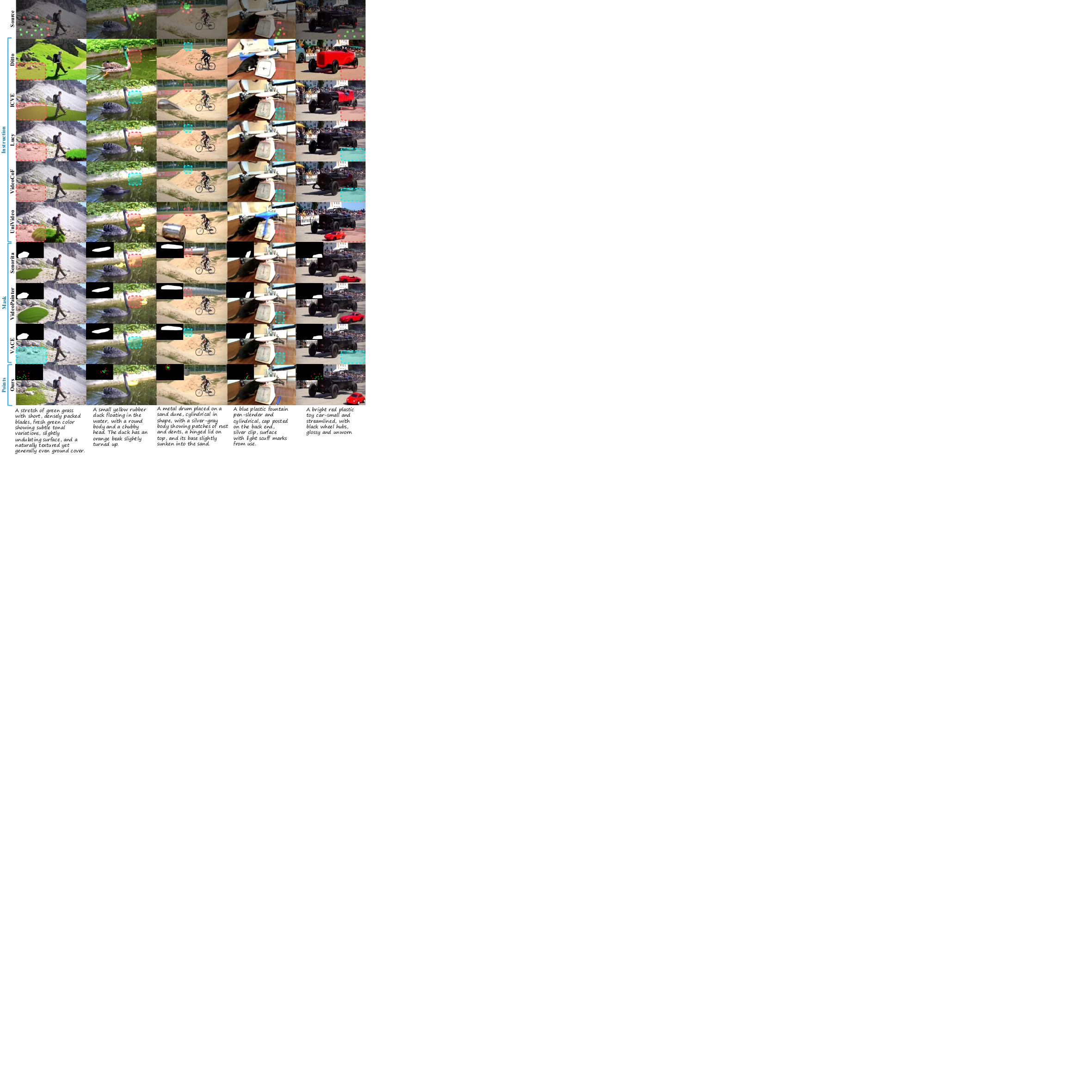} 
  }
    \vspace{-32pt}
  \caption{\textbf{Qualitative comparison of video object insertion.} Source videos (top) with a dark overlay contain sparse \textcolor{Green}{editable} and \textcolor{Red}{uneditable} points.
  To ensure a fair comparison, we adapt the input to handle these sparse signals: for \textit{instruction-based} models, QwenVL~\cite{qwen2.5-VL} translates source videos with markers into textual position prompts; for \textit{mask-based} models, the convex hull of the positive points across all keyframes is used as the input mask.  
  In contrast, \OurMethod\  directly consumes sparse point maps. 
  \textbf{Bounding boxes} highlight specific artifacts: \textcolor{red}{incorrect editing or wrong placement} and \textcolor{cyan}{insertion failure or unintended removal}.  
  Our method outperforms other methods in both photorealism and localization accuracy.
  }
  \label{fig:comparison}
\end{figure*}
\paragraph{\textbf{Qualitative Comparisons.}}
Fig.~\ref{fig:comparison} presents a qualitative comparison between our approach and state-of-the-art video editing methods. Because the initial frames of VideoPainter and Senorita are generated by external image editing models, which do not fully reflect the core video editing capabilities of these frameworks, we selected random middle frames for a more objective comparison.
The top row displays the original input frames with a black overlay, where green and red \textbf{points} indicate editable and non-editable regions. \textbf{Bounding boxes} highlight specific artifacts: red boxes denote misplacement, while cyan boxes indicate insertion failures.
Despite being provided with location instructions, current instruction-based methods lack sufficient positioning accuracy.Especially in scenarios with large regions of overall similar appearance (e.g., the rocks in column 1), these models fail to accurately localize the target location and have difficulty interpreting directional prompts such as \textit{left} or \textit{right}. Furthermore, the results of Ditto, ICVE, and Lucy are highly unstable, sometimes failing to insert the object. Additionally, VideoCoF even erroneously removes the swan’s head (column 2) and the black car’s front wheel (column 5).

Since mask-based methods require per-frame annotations, we use the convex hull mask of the point ranges as a proxy, which is inherently imprecise. In these cases, Senorita and VideoPainter tend to fill the entire mask, while VACE consistently fails to add any objects when handling such irregular, non-semantic masks. In contrast, our method outperforms existing approaches in bothvisual fidelity and localization accuracy, demonstrating robust object insertion even on DAVIS testset with significant camera motion.

\subsection{MLLM Evaluation}
Tab.~\ref{tab:mllm} reports the quantitative evaluation of editing quality on \textit{PointBench} using Gemini3-Pro\cite{team2024gemini}. Overall, our method achieves strong performance across metrics such as object insertion success, semantic alignment, and visual quality, outperforming most baselines. While Video Painter attains comparable results, it relies on a significantly larger model size (5B + 12B vs. our 1.3B) and requires additional first-frame image editing; in contrast, our method is more lightweight and versatile. 
Moreover, VLM-based models struggle to determine whether the inserted objects are placed at the correct locations with respect to the given control conditions. Consequently, our superiority in this aspect is more clearly reflected by the Acc$_{\text{pos}}$ and Acc$_{\text{neg}}$ scores in Tab.~\ref{tab:benchmark}.

\subsection{Ablation Analysis and Hyperparameter Tuning}
We perform ablation studies and hyperparameter tuning to assess method components, point densities, and point sizes.

\begin{table}[t]
\centering
\small
\caption{
\textbf{Ablation Studies on \OurMethod\  Method Components.}  
We perform ablation studies on the DAVIS subset of \Benchmark. In this context, \( \mathcal{L}_{\text{pa}} \) denotes the point-aware control loss, and \( \mathcal{L}_{\text{etd}} \) represents the explicit teacher distillation loss. 
}

\scalebox{1}{
\setlength{\tabcolsep}{1.2mm}{
\begin{tabular}{cc|cc|ccc|cc}
\toprule
    &
    & $\textbf{Acc}_{\textbf{\textrm{pos}}}\uparrow$ 
    & $\textbf{Acc}_{\textbf{\textrm{neg}}}\uparrow$
    & $\textbf{MSE}\downarrow$ 
    & $\textbf{PSNR}\uparrow$ 
    & $\textbf{LPIPS}_{\times 100}\downarrow$ 
    & $\textbf{CLIP BG}\uparrow$ 
    & $\textbf{CLIP TA}\uparrow$ 
    \\
\midrule
    \multicolumn{2}{c|}{\textbf{\xspace w/{o} $\mathcal{L}_{\textrm {pa}}$} }
    & 74.36 & \textcolor{Red}{98.04} & {97.87}  & {29.15} & {4.60} & \best{25.80} & \best{23.73}  \\

    \multicolumn{2}{c|}{\textbf{\xspace w/\phantom{o} $\mathcal{L}_{\textrm {pa}}$} }
    & \textcolor{Red}{78.26} & {97.89} & \textcolor{Red}{95.24}  & \textcolor{Red}{29.38} & \textcolor{Red}{4.55} & 25.73 & 23.70  \\

\midrule

    % & -- & -- & -- & -- & -- & -- & -- & -- & -- \\
    \multicolumn{2}{c|}{\textbf{\xspace w/{o} $\mathcal{L}_{\textrm {etd}}$} }
    & {78.93} & {96.35} & {101.08}  & {29.10}  & {4.63} & {25.68} & \textcolor{Red}{23.73}  \\

\multicolumn{2}{c|}{\textbf{\xspace w/\phantom{o} $\mathcal{L}_{\textrm {etd}}$}}     
    & \textcolor{Red}{79.77} & \textcolor{Red}{97.33} & \textcolor{Red}{94.74} & \textcolor{Red}{29.49}  & \textcolor{Red}{4.51} & \textcolor{Red}{25.74} & {23.64}  \\
\bottomrule
\end{tabular}
}}
\label{tab:abla_method}
\end{table}

\paragraph{\textbf{Effects of Explicit Teacher Distillation and Point-Aware Enhancement.}}
We conduct ablation studies on the effects of the point-aware control loss $\mathcal{L}_{\text{pa}}$
and the distillation loss $\mathcal{L}_{\text{etd}}$, as shown in Tab.~\ref{tab:abla_method}.
Introducing the point-aware loss $\mathcal{L}_{\text{pa}}$ leads to a clear improvement in control accuracy,
with the positive control accuracy Acc$_{\text{pos}}$ increasing from 74.36\% to 78.26\%.
These results indicate that $\mathcal{L}_{\text{pa}}$ effectively enforces point-level constraints, thereby improving the model's control over object location and structure.

By comparing settings with and without the explicit teacher distillation loss $\mathcal{L}_{\text{etd}}$,
we observe additional gains in both control accuracy and generation quality when teacher guidance is introduced.
Specifically, Acc$_{\text{pos}}$/Acc$_{\text{neg}}$ improve by nearly 1\%,
and PSNR increases by 0.3, while semantic alignment remains stable.
These results demonstrate that explicit teacher distillation provides complementary supervision, contributing to improved overall generation quality.

\paragraph{\textbf{Effects of Point Density.}}
In Tab.~\ref{tab:abla_ratio}, we conduct an ablation study on different control  settings, including sampling only at the first frame (First frame), sampling at keyframes with a fixed point density (Fixed density), and sampling at keyframes with varying point densities (Variable density).

Sampling points only from the first frame leads to a significant performance degradation, with $\text{Acc}_{\text{pos}}$ dropping by 7.8\%. 
This is because the control signals are too sparse, causing the model to fail to insert objects in subsequent frames, which leads to unstable object generation and temporal flickering.
In contrast, sampling at keyframes helps mitigate this issue. Among three setting, the various sampling ratio achieve the best overall performance, with $Acc_{\text{pos}}$ and $Acc_{\text{neg}}$ increasing by 1.42\% and 0.53\%, respectively. These results demonstrate the model’s strong generalization capability across varying point densities.

\begin{table}[t]
\centering
\small
\caption{
\textbf{Ablation Study on Different Point Density.}  
In this context, \textbf{First frame} refers to applying control signals only to the first frame. \textbf{Fixed density} refers to applying control signals at specified keyframes with a fixed spatial sampling density. \textbf{Variable density} refers to applying control signals at specified keyframes with varying spatial sampling densities.
}
\scalebox{1}{
\setlength{\tabcolsep}{1.2mm}{
\begin{tabular}{l|cc|ccc|cc}
\toprule
    & $\textbf{Acc}_{\textbf{\textrm{pos}}}\uparrow$ 
    & $\textbf{Acc}_{\textbf{\textrm{neg}}}\uparrow$
    & $\textbf{MSE}\downarrow$ 
    & $\textbf{PSNR}\uparrow$ 
    & $\textbf{LPIPS}_{\times 100}\downarrow$ 
    & $\textbf{CLIP BG}\uparrow$ 
    & $\textbf{CLIP TA}\uparrow$ \\
\midrule
\textbf{\xspace First frame} 
    & 65.14 & 97.39 & \textcolor{Red}{93.60}  & \textcolor{Red}{29.46}  & \textcolor{Red}{4.57} & \textcolor{Blue}{25.60} & \textcolor{Blue}{23.64}  \\

\textbf{\xspace Fixed ratio}    
    & \textcolor{Blue}{72.94} & \textcolor{Blue}{97.51} & \textcolor{Blue}{95.82}  & \textcolor{Blue}{29.29}  & \textcolor{Red}{4.57} & 25.52 & 23.52 \\
    
\textbf{\xspace Various ratio}    
    & \textcolor{Red}{74.36} & \textcolor{Red}{98.04} & 97.87  & 29.15  & \textcolor{Blue}{4.60} & \textcolor{Red}{25.80} & \textcolor{Red}{23.73}  \\
\bottomrule
\end{tabular}
}}
\label{tab:abla_ratio}
\end{table}

\begin{table}[t]
\centering
\small
\caption{
\textbf{Ablation Study on Different Point Size.} 
Results are more effective when the point size corresponds to the receptive field pixels of one visual token.
}
\scalebox{1}{
\setlength{\tabcolsep}{1.2mm}{
\begin{tabular}{l|cc|ccc|cc}
\toprule 
\textbf{Grid Size} & $\textbf{Acc}_{\textbf{\textrm{pos}}}\uparrow$ 
& $\textbf{Acc}_{\textbf{\textrm{neg}}}\uparrow$ 
& $\textbf{MSE}\downarrow$ 
& $\textbf{PSNR}\uparrow$ 
& $\textbf{LPIPS}_{\times 100}\downarrow$ 
& $\textbf{CLIP BG}\uparrow$ 
& $\textbf{CLIP TA}\uparrow$\\
\midrule
\textbf{\xspace 2} 
& 57.51 
& 94.40 
& \second{98.11} 
& 29.26 
& 4.63 
& 25.63 
& 23.68\\

\textbf{\xspace 6} 
& 70.28 
& 96.28 
& 99.89 
& \second{29.40} 
& \second{4.61} 
& 25.60 
& 23.58\\

\textbf{\xspace 10} 
& \best{72.78} 
& 95.32 
& 108.82 
& 29.10 
& 4.69 
& \second{25.68} 
& \best{23.73}\\

\textbf{\xspace 20}    
& \second{70.98} 
& \best{96.63} 
& 105.23 
& 29.23 
& 4.71 
& \best{25.69} 
& \second{23.71}\\

\textbf{\xspace 30}    
& 56.26 
& \second{96.43} 
& \best{92.81} 
& \best{29.49} 
& \best{4.59} 
& 25.59 
& 23.61 \\

\bottomrule
\end{tabular}
}}
\label{tab:abla_size}
\end{table}

\paragraph{\textbf{Effects of Point Size.}}
In Tab.~\ref{tab:abla_size}, we study the impact of different point sizes, specifically 2, 6, 10, 20, and 30. The results show that both excessively small and excessively large point sizes lead to a noticeable drop in accuracy. In particular, using a very extreme point sizes of 2 and 30 results in a 15.72\% and 16.52\% drops in $\text{Acc}_{\text{pos}}$, respectively. This is because small point sizes produce weak control, which are often ignored by the model and lead to object insertion failures in some cases, whereas large point sizes reduce control flexibility. In contrast, point sizes of 10 and 20 achieve the highest and second-highest $\text{Acc}_{\text{pos}}$ scores, respectively. Overall, these results indicate that point sizes in the range of 10–20 provide a better trade-off between control strength and flexibility, which aligns with the scale of the model’s $16 \times 16$ patch downsampling factor.

\section{Conclusion and Discussion}\label{sec-5-con}

In this paper, we present \OurMethod, a point-based framework for user-friendly video object addition that introduces three key innovations:  unified representation integrating point and mask control inputs; a mask-guided object insertion model distilled into a point-guided model, enabling performance comparable to mask-based methods even with sparse point inputs; and a large-scale dataset comprising 1.3M video pairs, along with a benchmark designed to evaluate the effectiveness of current object addition models. Extensive experiments demonstrate that \OurMethod\  achieves state-of-the-art results in both precise control and visual quality. Nevertheless, \OurMethod\  also has limitations: when global prompts are applied, object insertion capability slightly degrades due to the imprecise description; and the editing quality is constrained by the base model’s relatively small size (1.3B parameters). We anticipate that scaling to larger models will enhance performance in complex scenes and further improve visual quality.
\begin{figure*}[htbp]
  \centering
  \adjustbox{trim={0.0\width} {0.2\height} {0.434\width} {0}, clip, width=0.90\textwidth}{
    \includegraphics{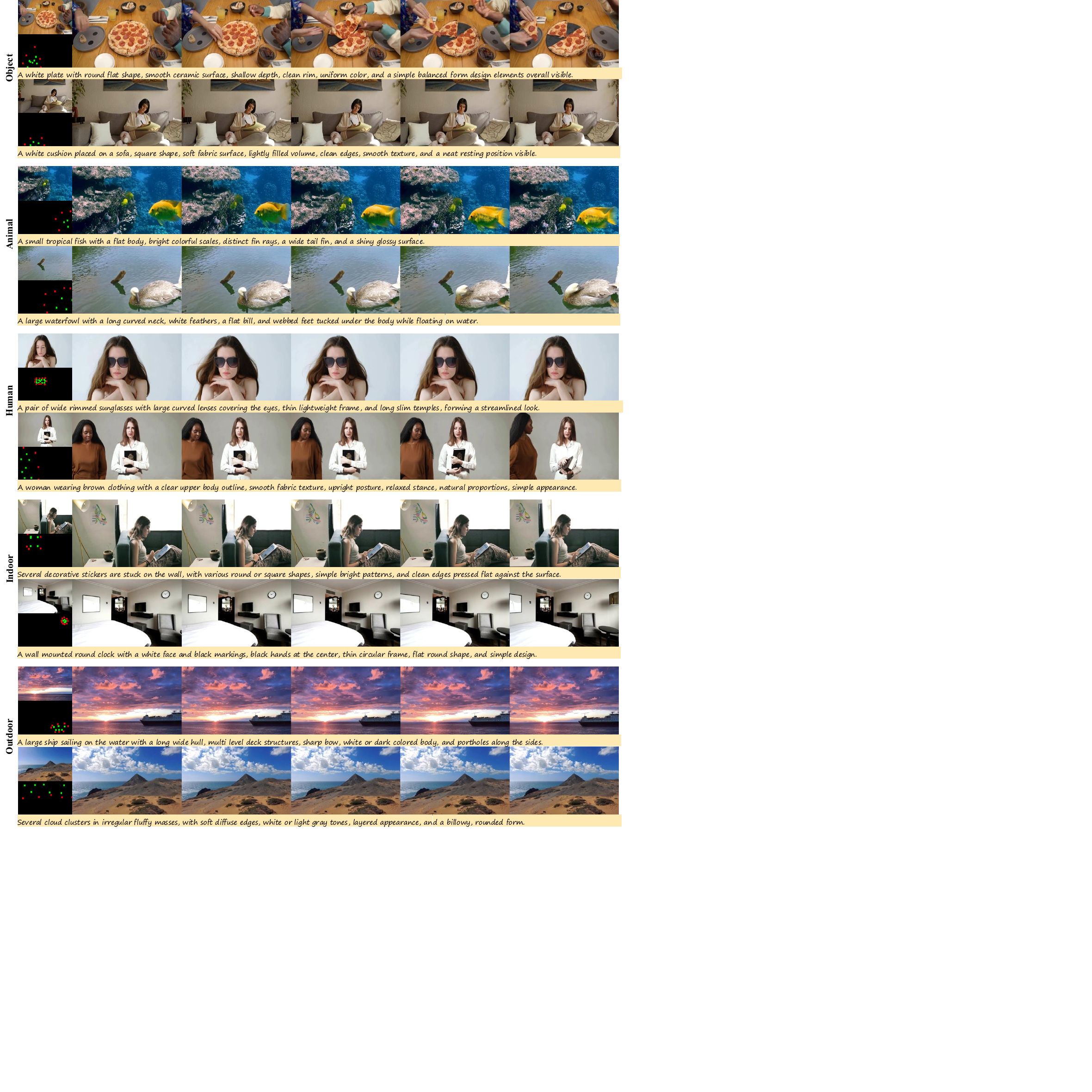} 
  }
   \vspace{-30pt}
    \captionsetup{justification=centering}
  \caption{\textbf{More video object insertion results.} 
  }
  \label{fig:comparison_2}
\end{figure*}

\begin{figure*}[htbp]
  \centering
  \adjustbox{trim={0.0\width} {0.222\height} {0.666\width} {0}, clip, width=0.90\textwidth}{
    \includegraphics{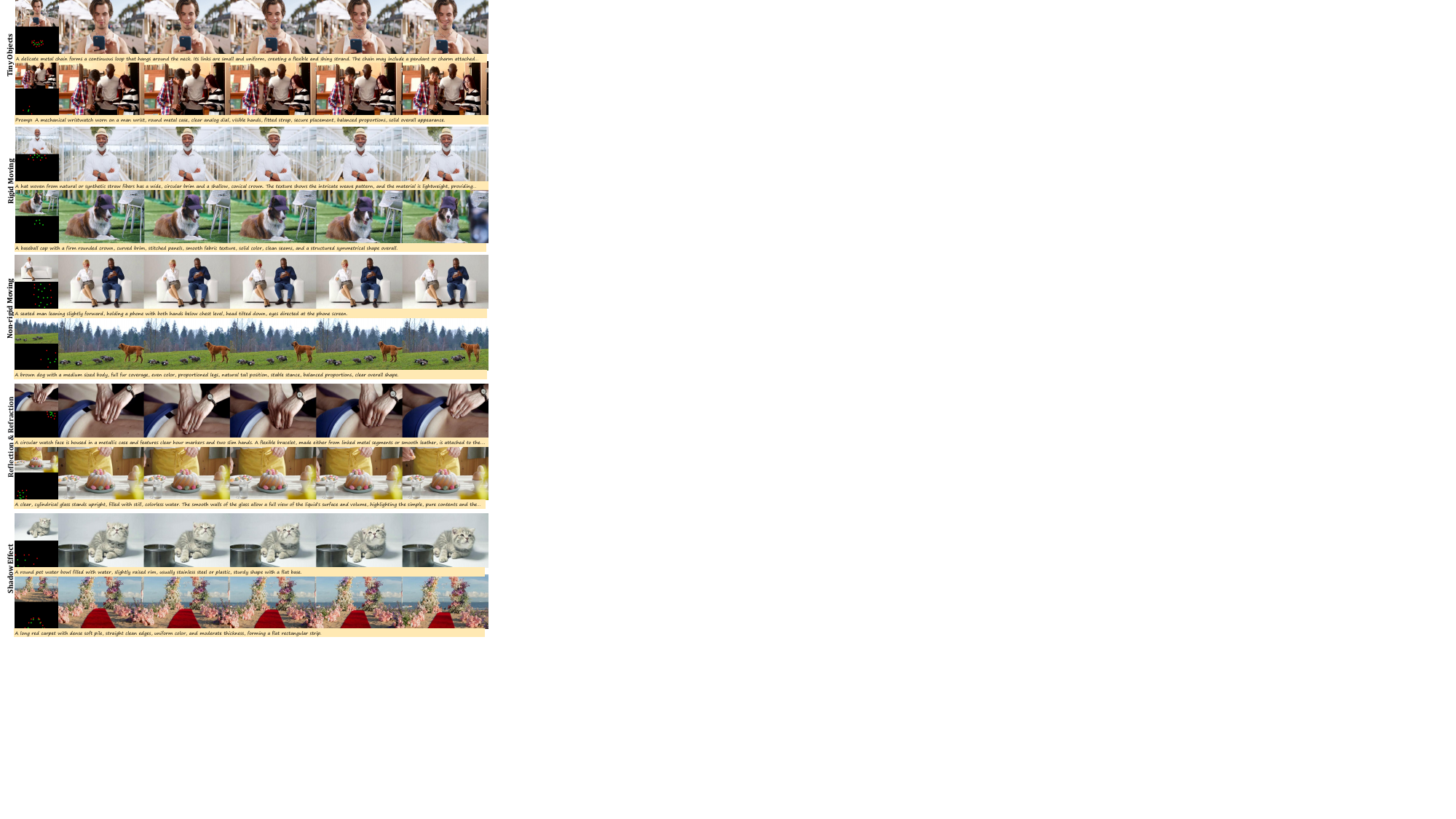} 
  }
   \vspace{-2pt}
    \captionsetup{justification=centering}
  \caption{\textbf{More video object insertion results.} }
  \label{fig:vis3}
\end{figure*}

\newpage

\bibliographystyle{ACM-Reference-Format}
\bibliography{paper}

% Appendix
\newpage
\clearpage

\appendix

\section{User study}

To evaluate the perceptual alignment of our method with human judgment, we conduct a formal user study comparing our approach against four state-of-the-art baselines: two instruction-based and two mask-based methods. 
We recruit 21 participants to assess the editing results. For each test case, users are presented with the original and edited videos and are asked to rate the performance on a 5-point Likert scale across three dimensions: 
\begin{itemize}[leftmargin=*]
\item Generative Quality: This metric assesses the realism of the inserted objects, their alignment with the text prompts (in terms of appearance and motion), and the presence of temporal artifacts such as flickering or jitter. 
\item Background Preservation: We evaluate the model’s ability to maintain the integrity of the original scene, ensuring that non-target regions remain unmodified and the overall visual style stays consistent. 
\item Spatial Accuracy: This criterion focuses on the plausibility of the spatial layout, including the accuracy of object positioning, scale proportions, and the correctness of occlusion relationships with the environment.
\end{itemize}

\begin{table}[htp]
\centering
\small
\caption{\textbf{User study}.
We conduct a user study comparing our method with two mask-based and two instruction-based state-of-the-art video editing models. \best{Red} and \second{blue} indicate the best and second-best performance.
}

\begin{tabular}{l|cccc}
\toprule
\multirow{2}{*}{\textbf{Method}}
& \textbf{Generative}
& \textbf{Background}
& \textbf{Spatial}
& \multirow{2}{*}{\textbf{Overall}} \\
& \textbf{Quality}
& \textbf{Preservation}
& \textbf{Accuracy}
& \\
\midrule
\textbf{Lucy}   & 2.152 & 2.852 & 2.204 & 2.403 \\
\textbf{UniVideo} & \second{2.652} & 2.809 & 2.357 & 2.606 \\
\textbf{Video Painter}    & 2.619 & \second{2.953} & \second{2.847} & \second{2.807} \\
\textbf{Vace}   & 2.000 & 2.952 & 2.333 & 2.429 \\
\midrule
\textbf{Ours}   & \best{3.858} & \best{4.181} & \best{4.157} & \best{4.065} \\
\bottomrule
\end{tabular}
\label{tab:user_study}
\end{table}

\section{Failure Case}

% \begin{wrapfigure}{l}{.45\textwidth}
% \end{wrapfigure}
\begin{figure}[h]
    \centering
    \scalebox{1.8}{
        \includegraphics{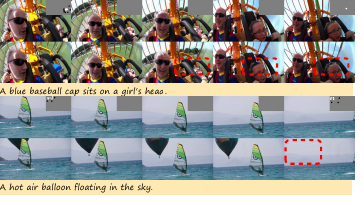} 
    }
    % \captionsetup{justification=centering}
    \caption{\textbf{Failure Case.} 
        When the scene involves high-speed motion or only the initial frame is annotated, the inserted object may suddenly disappear.
    }
    \label{fig:fail}
\end{figure}

While our model can insert static or dynamic objects into both static and moving scenes, failures may still occur in videos characterized by large-scale motion or significant camera shake. Additionally, insufficiently dense annotations can lead to insertion failures. For instance, if a point is provided only in one frame, the inserted object might disappear in subsequent frames. Addressing these limitations is reserved for future work. Visual examples of these failure cases are shown in Fig. \ref{fig:fail}.

\section{Ablation Study on Sparse Point and Dense Mask Input}

The difficulty of the object insertion task varies significantly depending on the type of control input provided. To investigate this, we trained two separate models, one guided exclusively by \textbf{sparse points} and the other by \textbf{dense mask}, and evaluated their performance after an equal number of training iterations.

As illustrated in Figure \ref{fig:mask-vs-point}, precise frame-by-frame mask guidance provides the model with explicit localization and boundary information, thereby significantly lowering the barrier for successful object insertion. In contrast, achieving precise insertion under sparse point conditions presents a much greater challenge. In the absence of additional boundary information, the model must independently determine the editing contours. After same training steps, the model struggles to establish a robust mapping between semantic concepts and the generation of object silhouettes. This often leads to distorted outlines, incomplete or misaligned boundaries, or failed generations resulting in blurred color patches.

The quantitative metrics presented in Tab.~\ref{tab:abla_condition} further substantiate the increased complexity of the point-guided task. In this setting, the model is required to synthesize the insertion position, the temporal video context, and the textual conditions simultaneously. The results indicate that the mask-based model outperforms the point-based model by 4.99 in PSNR and 1.65 in LPIPS, respectively, thanks to the clear object boundaries provided by the masks. The mask-guided approach consistently demonstrates superior performance across other metrics as well, further highlighting the inherent advantages of dense signals over sparse signals for fine-grained control. This observation inspired our development of a two-stage training strategy and a mask-to-point distillation method, which facilitates an effective transition from dense to sparse control signals.

\begin{table}[htp]
\centering
\small
\caption{
\textbf{Ablation Studies on Control Signal Settings.} 
We conduct experiments using points and masks as control during both training and inference. To ensure a fair comparison, all other experimental settings—including the dataset, architecture, and training iterations—are held constant. The results demonstrate that the mask-based method consistently outperforms the point-based counterpart in background preservation, semantic alignment, and temporal consistency. This performance gain is attributed to the explicit spatiotemporal boundaries and denser control inherent in masks.
}
\scalebox{0.9}{
    \setlength{\tabcolsep}{1.2mm}{
    \begin{tabular}{l|ccccc|cc|cc}
\toprule
    & $\textbf{MSE}\downarrow$ 
    & $\textbf{MAE}\downarrow$ 
    & $\textbf{PSNR}\uparrow$ 
    & $\textbf{SSIM}\uparrow$ 
    & $\textbf{LPIPS}_{\times 100}\downarrow$ 
    & $\textbf{CLIP BG}\uparrow$ 
    & $\textbf{CLIP TA}\uparrow$ 
    & $\textbf{CLIP TC}\uparrow$ 
    & $\textbf{EWarp}\downarrow$ 
    \\
\midrule
\textbf{\xspace Mask} 
    & \best{65.58} & \best{2.30} & \best{36.66} & \best{0.9714} & \best{1.79} & 24.28 &\best{23.38} & \best{99.34} & \best{0.35} \\

\textbf{\xspace Point}    
    & 133.41 & 5.23 & 31.67 & 0.9433 & 3.44 & \best{24.35} & 23.02 & 99.25 & 0.38\\

\bottomrule
\end{tabular}
}}
\label{tab:abla_condition}
\end{table}

\section{Model Parameter}
\label{supp:param}
In this section, we briefly describe the architectures and parameter sizes of different baselines. As summarized in Tab.~\ref{tab:benchmark}, we group existing methods into two categories: \textbf{instruction-based} video editing methods and \textbf{mask-based} video editing methods. The reported \emph{Model Size} corresponds to the parameters required at inference time. When a method relies on additional models (e.g., for multimodal understanding or inpainting/filling), we report the total size in the form of ``A+B''.

\paragraph{Instruction-based methods.}
\begin{itemize}
    \item \textbf{Ditto}~\cite{ditto_bai2025scaling} is built upon Wan 2.1-T2V-14B~\cite{wan2025} and trains an additional 0.2B LoRA for editing adaptation.
    \item \textbf{Lucy}~\cite{decartai2025lucyedit} is based on Wan 2.2-T2V-5B, resulting in a 5B-scale instruction-driven editing model.
    \item \textbf{ICVE}~\cite{icve_liao2025context} is built upon HunyuanVideo-T2V-13B~\cite{hunyuanvideo_kong2024hunyuanvideo}, forming a 13B instruction-based video editing model.
    \item \textbf{UniVideo}~\cite{univideo_wei2026univideounifiedunderstandinggeneration} adopts a HunyuanVideo backbone and integrates Qwen-VL-2.5-7B~\cite{qwen2.5-VL} for multimodal understanding, resulting in a total size of {13B+7B}.
    
    \item \textbf{VideoCoF}~\cite{videocof_yang2025unifiedvideoeditingtemporal} is based on Wan-2.1-T2V-14B with a model size of 14B.
\end{itemize}

\paragraph{Mask-based methods.}
\begin{itemize}
    \item \textbf{Senorita}~\cite{senorita_zi2025se} uses {CogVideoX-5B-I2V}~\cite{cogvideox_yang2024cogvideox} as the backbone and employs {FLUX.1-Fill [dev]~12B}~\cite{flux} for first-frame guided filling/editing, resulting in a total of {5B+12B}.
    \item \textbf{VideoPainter}~\cite{bian2025videopainter} adopts the same backbone and filling model as Senorita, hence also {5B+12B}.
    \item \textbf{Vace}~\cite{vace_jiang2025vace} adopts a context-adaptation architecture, built on {Wan 2.1-T2V-1.3B} with an additional {0.3B} adaptation module, yielding {1.3B+0.3B} in total.
\end{itemize}

Our method requires only a {1.3B} parameter model at inference time while achieving SOTA editing performance.

\section{MLLM Evaluation Accuracy}
To rigorously assess the video editing performance, we employ state-of-the-art Multimodal Large Language Models (MLLMs), specifically Gemini3-Pro \cite{team2024gemini} and GPT-5.2 \cite{openai_gpt5_2025}, as automated evaluators. To validate the reliability of these model-based metrics, we conduct a comparative analysis against human-annotated ground truth.

For the \textbf{Gemini3-Pro} evaluation, we randomly sampled 600 instances for human-model alignment verification. Each test case consists of an original-edited image pair, where a red bounding box defines the target Region of Interest (ROI). The models are tasked with identifying whether the requested object was successfully synthesized within the designated ROI. Gemini3-Pro achieves a high agreement rate of 97.2\% with human annotators. Qualitative analysis of the remaining 2.8\% discrepancies reveals that most errors stem from inherent semantic ambiguities at transitional boundaries—for instance, distinguishing forest elements from adjacent grassland. In such boundary regions, pre-existing visual contexts often confound the model's categorical judgment.

\vspace{4mm}
\begin{tcolorbox}[colback=white,colframe=black!75!white,title=Gemini Prompt, breakable]
You will see two images and an editing instruction: the first is the original image, and the second is the target image.

Analyze the provided information and answer the following questions:

\begin{enumerate}[leftmargin=*,topsep=0.5em,itemsep=0.5em]
    \item \textbf{Object Addition Detection:} Was a new object added inside the red bounding box?
    
    \item \textbf{Semantic Alignment Scoring:} Evaluate the semantic consistency between the object in the red box and the given prompt, Give a score from 1 to 5.
    
    \item \textbf{Visual Quality Scoring:} Evaluate the visual quality of the added object, considering edge blending, lighting and shadow consistency, texture realism, perspective alignment, and overall coherence with the background. Give a score from 1 to 5.
\end{enumerate}

Return your response strictly in the following JSON format:

\begin{lstlisting}[basicstyle=\ttfamily\small,breaklines=true,breakatwhitespace=true]
{
  "has_added_object_in_red_box": true/false,
  "text_alignment": {
    "score": 1-5,
    "comments": "Brief explanation of the semantic assessment"
  },
  "visual_quality": {
    "score": 1-5,
    "comments": "Brief explanation of the quality assessment"
  },
  "reasoning": "Concise justification for all answers above"
}
\end{lstlisting}

\vspace{1em}
Do not include any additional text outside the JSON block.

\end{tcolorbox}

\vspace{4mm}

We further evaluate \textbf{GPT-5.2} assessment accuracy using 200 randomly selected samples under the same task formulation. GPT-5.2 demonstrates a 96\% alignment with human judgment. A granular error analysis of the eight discrepant cases identifies one false negative and seven false positives. We observe that false positives primarily occur when the synthesized object is localized near but strictly outside the bounding box, suggesting a slight misalignment in the model's spatial reasoning relative to the explicit ROI constraints.

%The scoring prompt templates are provided below.

\vspace{4mm}
\begin{tcolorbox}[colback=white,colframe=black!75!white,title=GPT Prompt, breakable]
You will see two images and an editing instruction: the first is the original image, and the second is the target image.

Analyze the provided information and answer the following questions:

\begin{enumerate}[leftmargin=*,topsep=0.5em,itemsep=0.5em]
    \item \textbf{Object Addition Detection:} Was a new object added inside the red bounding box? 
    
    \item \textbf{Semantic Alignment Scoring:} Evaluate the semantic consistency between the object in the red box and the given prompt. Score from 1 (completely inconsistent) to 5 (matches the prompt). 
    
    \item \textbf{Visual Quality Scoring:} Evaluate the visual quality of the added object (if any), considering factors such as edge blending, lighting/shadow consistency, texture realism, perspective alignment, and overall coherence with the background. Provide a quality score from 1 to 5 (5 = photorealistic and seamless; 1 = obvious artifact or poor integration).
\end{enumerate}

Return your response strictly in the following JSON format:

\begin{lstlisting}[basicstyle=\ttfamily\small,breaklines=true,breakatwhitespace=true]
{
  "has_added_object_in_red_box": true/false,
  "text_alignment": {
    "score": 1-5,
    "comments": "Brief explanation of the semantic assessment"
  },
  "visual_quality": {
    "score": 1-5,
    "comments": "Brief explanation of the quality assessment"
  },
  "reasoning": "Concise justification for all answers above"
}
\end{lstlisting}

\vspace{1em}
Do not include any additional text outside the JSON block.

\end{tcolorbox}

\vspace{4mm}

\begin{figure*}[htbp]
  \centering
  \adjustbox{trim={0.0\width} {0.32\height} {0.4\width} {0}, clip, width=0.99\textwidth}{
    \includegraphics{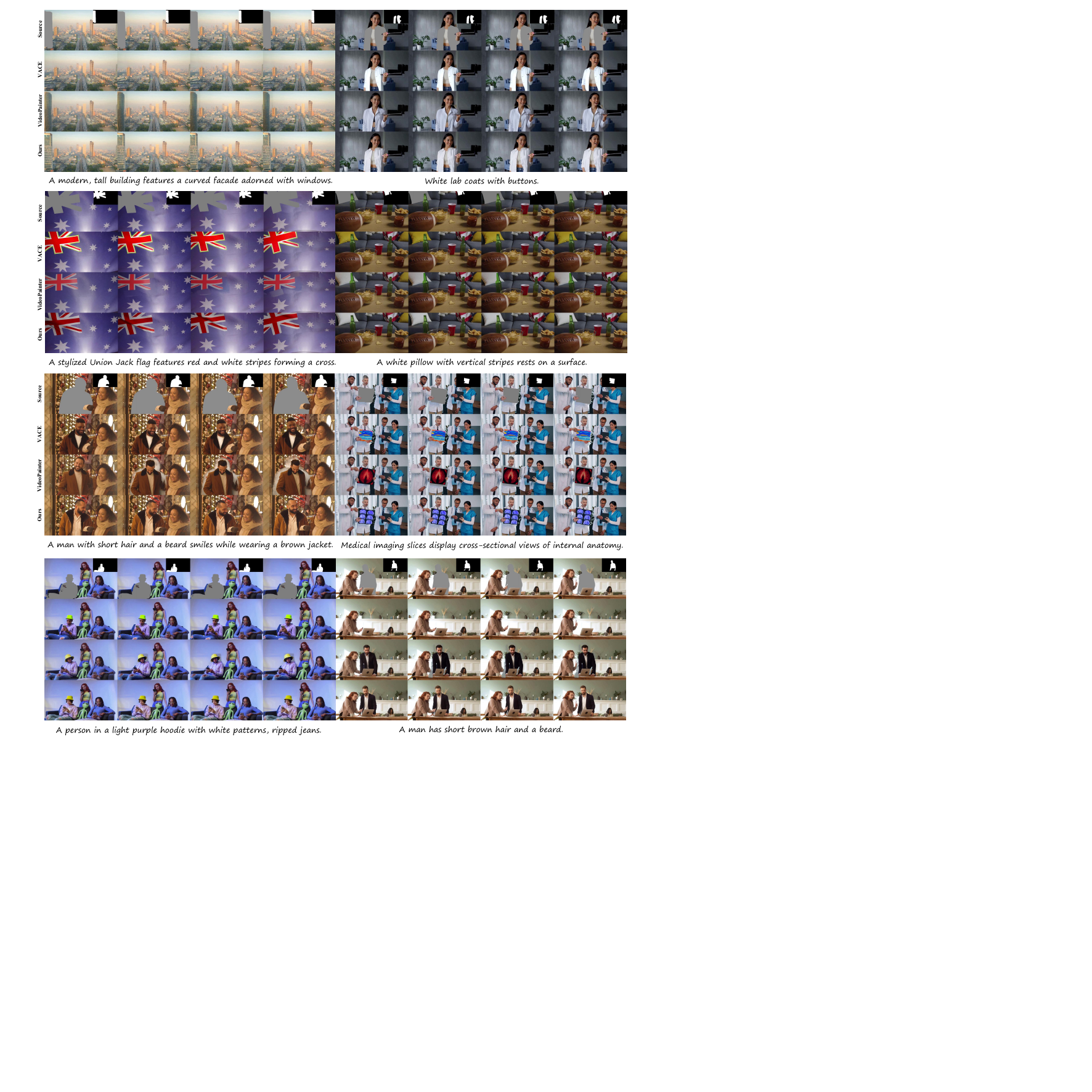}
  }
   \vspace{-2pt}
  \caption{\textbf{Mask-based Video Editing results.} 
    Using video and mask frames inputs, we compare the qualitative performance of VACE, VideoPainter, and our proposed method. While VACE occasionally fails to generate the target object and VideoPainter exhibits noticeable artifacts around object boundaries, our method consistently produces high-quality, semantically aligned editing results.
  }
  \label{fig:mask}
\end{figure*}

\end{document}